\pdfoutput=1

\documentclass[11pt]{article}

\usepackage[]{EMNLP2023}

\usepackage{times}
\usepackage{latexsym}

\usepackage[T1]{fontenc}

\usepackage[utf8]{inputenc}

\usepackage{microtype}

\usepackage{inconsolata}
\usepackage{adjustbox}
\usepackage{enumitem}
\usepackage[normalem]{ulem}

\newcommand{\ourmethod}{ATE-D}
\newcommand{\ratedistort}{TE-D}

\title{Debiasing Multimodal Models via Causal Information Minimization}

\author{Vaidehi Patil \ \ \ \ \ \ \ \
Adyasha Maharana \ \ \ \ \ \ \ \
Mohit Bansal \\
UNC Chapel Hill \\
\small\texttt{\{vaidehi, adyasha, mbansal\}@cs.unc.edu}\\
}

\begin{document}
\maketitle
\begin{abstract}

Most existing debiasing methods for multimodal models, including causal intervention and inference methods, utilize approximate heuristics to represent the biases, such as shallow features from early stages of training or unimodal features for multimodal tasks like VQA, etc., which may not be accurate. In this paper, we study bias arising from confounders in a causal graph for multimodal data and examine a novel approach that leverages causally-motivated information minimization to learn the confounder representations. Robust predictive features contain diverse information that helps a model generalize to out-of-distribution data. Hence, minimizing the information content of features obtained from a pretrained biased model helps learn the simplest predictive features that capture the underlying data distribution. We treat these features as confounder representations and use them via methods motivated by causal theory to remove bias from models. We find that the learned confounder representations indeed capture dataset biases, and the proposed debiasing methods improve out-of-distribution (OOD) performance on multiple multimodal datasets without sacrificing in-distribution performance. Additionally, we introduce a novel metric to quantify the sufficiency of spurious features in models' predictions that further demonstrates the effectiveness of our proposed methods.\footnote{Our code is available at: \url{https://github.com/Vaidehi99/CausalInfoMin}}

\end{abstract}

\section{Introduction}

\label{sec:intro}

\begin{figure}
    \centering
    \includegraphics[width=0.49\textwidth]{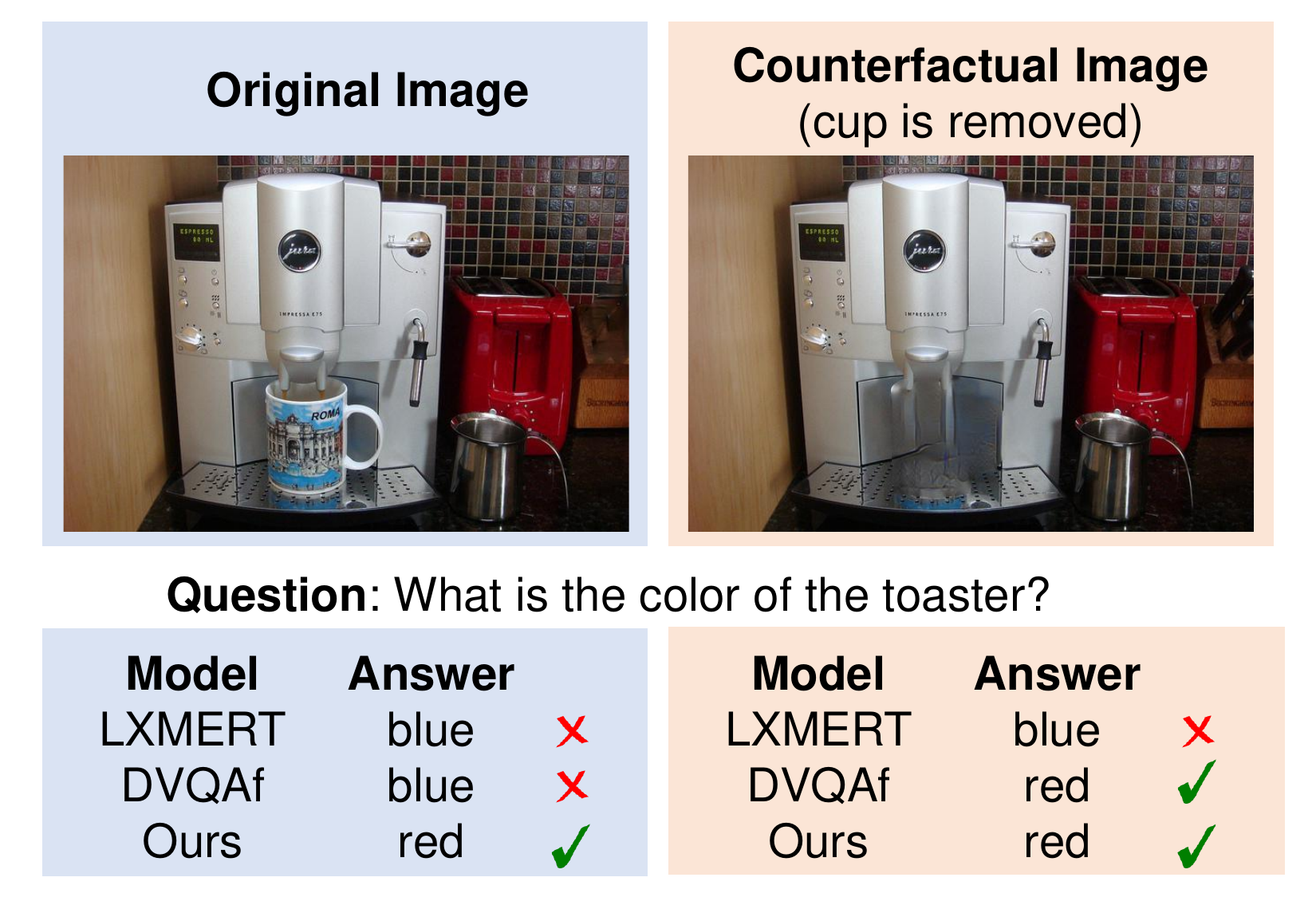}
    \caption{Multimodal models tend to rely on spurious correlations in the dataset to answer a question. Existing methods remove unimodal biases, whereas our method removes biases arising from cross-modal interactions as well and is more invariant to irrelevant features (e.g., the coffee mug) in this example.}
    \label{fig:intro}
    \vspace{-10pt}
\end{figure}

The success of multimodal models in various tasks has been attributed to their ability to rely on spurious correlations (or biases) present in the training data \cite{10.1007/978-3-319-46484-8_44, agrawal-etal-2016-analyzing, zhang2016yin, goyal2017making}. An example of image bias in VQA is when the model tends to look at prominent objects in the image rather than focusing on the object about which the question is asked \cite{wen2021debiased} (see Fig.~\ref{fig:intro}). These models leverage such biases to perform well on in-distribution (ID) evaluation data \cite{vqa-cp}. However, their poor performance on out-of-distribution data reveals that they merely rely on superficial features rather than capturing the true causal relationships between inputs and targets.

Existing methods attempt to diminish a model’s reliance on these shortcuts by taking one or both of two primary strategies: (a) by balancing the sample groups with and without spurious correlation, e.g. via data augmentation \cite{gokhale-etal-2020-mutant} or sample synthesis \cite{chen2020counterfactual, DBLP:conf/eccv/0016Z022, Kolling_2022_WACV}, and (b) by explicitly eliminating the impact of spurious correlations during model training or inference \cite{huang2022deconfounded, lin2022causal,pan2022causal}. In the former approach, the identification of the unique set of spurious correlations in each sample becomes essential to curate augmented samples for achieving balance. Consequently, approaches that alleviate biases in features or predictions, independent of the availability of non-spurious data, are more desirable. Such methods also offer the additional advantage of being agnostic to the specific dataset and task at hand.

\begin{figure}
    \centering
    \includegraphics[width=0.42\textwidth]{
    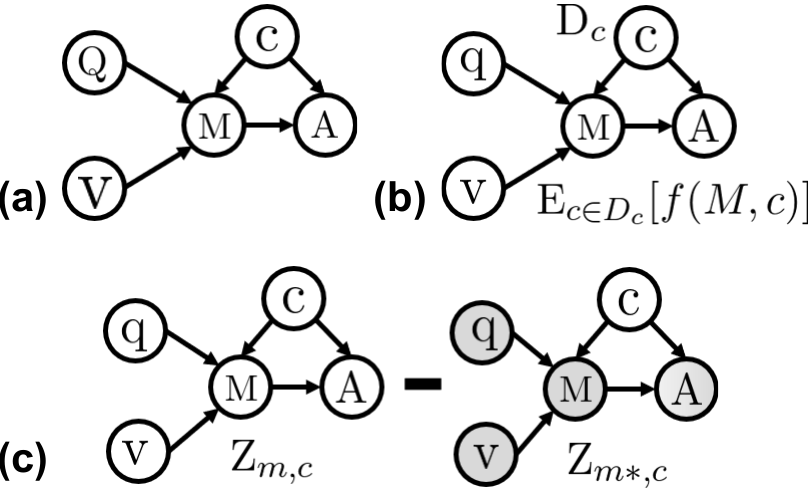}
    \caption{Demonstration of (a) our proposed causal graph for multimodal tasks, (b) Average Treatment Effect (ATE), and (c) Total Effect (TE) on (a). Values in grey indicate the `no-treatment' condition.}
    \label{fig:causal_graph}
    \vspace{-10pt}
\end{figure}

Recent research on debiasing models has emphasized the significance of causal theory \cite{zhang-etal-2021-de, liu2022contextual, bahadori2020debiasing} i.e., many spurious correlations originate from confounding variables that induce non-causal dependencies between inputs and labels \cite{pearl2000models}. However, effectively identifying and representing biases that undermine prediction accuracy remains a challenging task. Previous studies on multimodal models have utilized image features from early training stages as contextual biases for multi-label image classification \cite{liu2022contextual}, or introduced unimodal training branches to mitigate spurious correlations in Visual Question Answering (VQA) \cite{niu2021counterfactual}. Moreover, these approaches overlook biases stemming from multimodal interactions within their causal graphs. Hence, in this work, we represent the bias as confounder variables that have a direct causal effect on multimodal features and the corresponding predictions (see Fig.~\ref{fig:causal_graph}(a)). Spurious correlations represent the simplest predictive features that explain biased datasets \cite{geirhos2020shortcut}, thereby making them easily learnable by machine learning models under limited representation capacity \cite{yang2022chroma}. We capitalize on this notion to study a novel framework that combines information theory and causal graphs to learn confounder representations capable of capturing spurious features. We examine two approaches to learn the confounder representations by imposing information loss on biased multimodal features i.e., (a) \textit{latent variable modeling} using a generative model and (b) \textit{rate-distortion} minimization \cite{shannon1948mathematical}. Subsequently, we utilize these confounders in our proposed debiasing methods, namely \ourmethod{} and \ratedistort{}, leveraging the concepts of \textit{average treatment effect} \cite{glymour2016causal} and \textit{total effect} \cite{pearl2022direct} causal mechanisms, respectively.

In \ourmethod{}, we employ an autoencoder to reconstruct the biased features. The autoencoder projects these features into a lower-dimensional latent space, capturing latent features that act as substitutes for unobserved confounders \cite{huang2022deconfounded}. By clustering the learned confounder representations across the dataset, we construct a dictionary of confounders. We subsequently perform backdoor adjustment based on the average treatment effect, utilizing feature reweighting \cite{kirichenko2022last}. In \ratedistort{}, we leverage the \textit{rate-distortion function}, which controls the number of bits required to encode a set of vector representations \cite{chowdhury-chaturvedi-2022-learning}. We minimize the rate-distortion function for a non-linear projection of the features extracted from a biased pretrained model, while simultaneously minimizing the cross-entropy loss of predicting from these projected features. This results in the loss of diverse information from the features and the retention of simple features that are also maximally predictive of the biased dataset. We treat these features as the confounder representations that stem from spurious correlations in the dataset and compute the (unbiased) \textit{total effect} of the input by taking the difference between the biased feature and its respective confounder.

We evaluate the proposed methods on several multimodal tasks and along multiple dimensions i.e., in-distribution and out-of-distribution performance, efficiency, and robustness. Results show that these methods not only outperform baseline models with lower training overhead but also yield additional gains on top of unimodal debiasing methods. In this work, we demonstrate the presence of multimodal biases and the need for multimodal debiasing along with the potential of confounder modeling via information loss in causal multimodal debiasing. Our contributions are as follows:

\begin{itemize}[leftmargin=*]
\setlength\itemsep{0em}
\item We present two methods, \ratedistort{} and \ourmethod{}, that leverage causally-motivated information loss to learn confounder representations from biased features and utilize them to debias models.
\item Our methods remove multimodal biases and yield up to 2.2\% and 2.5\% gains over LXMERT \cite{tan-bansal-2019-lxmert}, on VQA-CP and GQA-OOD \cite{kervadec2021roses} datasets respectively, and 0.7\% gains on top of unimodal debiasing \cite{wen2021debiased}. Importantly, our methods exhibit superior parameter efficiency and reduced training time compared to existing debiasing methods.
\item We propose a sufficiency score ($\lambda$) for quantifying the reliance of models on spurious features. Results show that our methods improve robustness to spurious correlations in the dataset.
\item We analyze the confounders learnt in \ourmethod{}, \ratedistort{} and show that they encode dataset biases.

\end{itemize}

\section{Related Work}

\paragraph{Data Augmentation.} Balancing data \cite{DBLP:conf/cvpr/ZhangGSBP16} can involve training a generative model for sample synthesis \cite{agarwal2020towards, sauer2020counterfactual}, designing suitable data selection heuristics \cite{chen2020counterfactual}, or curating balanced/counterfactual samples \cite{goyal2017making, gokhale-etal-2020-mutant, 9706896}. Human explanations can be used as additional training signals to promote reasoning \cite{yingvisfis, wu2019self, selvaraju2019taking}. We debias models using existing biased data.

\paragraph{Inductive Bias in Model Architecture.} \citet{vqa-cp} explicitly design inductive biases to prevent the model from relying on training priors. \citet{clark2019don, cadene2019rubi, ramakrishnan2018overcoming} rely on a separate QA branch to weaken the language prior in VQA models via adversarial or multi-task learning. \citet{wen2021debiased} use a contrastive loss to remove unimodal biases for VQA. \citet{peyrard-etal-2022-invariant} discover invariant correlations in data across different training distributions to enable generalization.

\paragraph{Inductive Bias for Modeling Confounders.} 
\citet{kallus2018causal} recover latent confounders via low-rank matrix factorization and \citet{sen2017contextual} utilize low-dimensional variables for encoding confounders. We use low-dimensional features to limit representational capacity for encoding confounders in multimodal data.

\paragraph{Causal Perspective.} \citet{lin2022causal} use causal intervention through backdoor adjustment \cite{glymour2016causal} to disentangle the biases for unsupervised salient object detection. \citet{huang2022deconfounded} use ATE to debias referring expression models. \citet{niu2021counterfactual} compute the Total Indirect Effect (TIE) of the multimodal branch to omit the influence of unimodal branches. \citet{veitch2021counterfactual} formalize counterfactual invariance and its relation to OOD performance. \citet{liu2022contextual} use features from early training as confounders and compute the Total Direct Effect (TDE) for multi-label image classification. We combine information theory and causal theory to learn confounders from biased representations and use them via ATE and TE causal mechanisms to debias a model.

\section{Causal Theory Preliminaries}
\vspace{-5pt}
\label{sec:prelim}
In this section, we discuss our proposed causal graph for multimodal tasks and the two causal mechanisms relevant to our debiasing methods.

\paragraph{Causal Graph.} Causal graphs are directed acyclic graphs $\mathcal{G}=\{\mathcal{V}, \mathcal{E}\}$ where the edges $\mathcal{E}$ are used to represent causal relationships between random variables $\mathcal{V}$. When the variable $\mathbf{Q}$ has an \textit{indirect effect} on $\mathbf{A}$ through a variable $\mathbf{M}$ i.e. $\mathbf{Q}\rightarrow\mathbf{M}\rightarrow\mathbf{A}$, the variable $\mathbf{M}$ is said to be a \textit{mediator} in the causal graph (see Fig.~\ref{fig:causal_graph}(a)). If a variable $\mathbf{C}$ has a \textit{direct causal effect} on both $\mathbf{M}$ and $\mathbf{A}$, it is said to be a \textit{confounder}.

\paragraph{Causal Perspective for Multimodal Tasks.}

Multimodal models for tasks combining vision ($V$) and language ($Q$) often face the challenge of confounding variables, which introduce spurious features. Current approaches rooted in causal theory aim to mitigate direct unimodal effects. However, a VQA example (Fig.~\ref{fig:intro}) highlights a limitation: models trained predominantly on centrally located objects struggle with queries about obscured object colors. Existing causal graphs for multimodal tasks fail to account for spurious correlations arising from such multimodal interactions. To address this, we propose a confounder $\mathbf{C}$ that influences both the mediator $\mathbf{M}$ and the answer $\mathbf{A}$ (Fig.\ref{fig:causal_graph}(a)). By modeling biases encoded in multimodal features as confounder $\mathbf{C}$, we can eliminate biases using causal intervention.

In order to debias VQA models, we adopt two causal mechanisms i.e., the Average Treatment Effect (ATE) and Total Effect (TE), which essentially refer to the same quantity but differ in how they deal with the confounder \cite{vanderweele2015explanation, tang2020unbiased}. In ATE, $C$ is treated as a distribution, and $c$ is sampled by assuming implicit causal association with the treatment $M=m$. In TE, $c$ has an explicit causal association with the treatment $M=m$ in each sample. We explore both in our work and discuss their theories below.

\begin{figure*}[ht]
\begin{center}
  \includegraphics[width=0.85\linewidth]{
  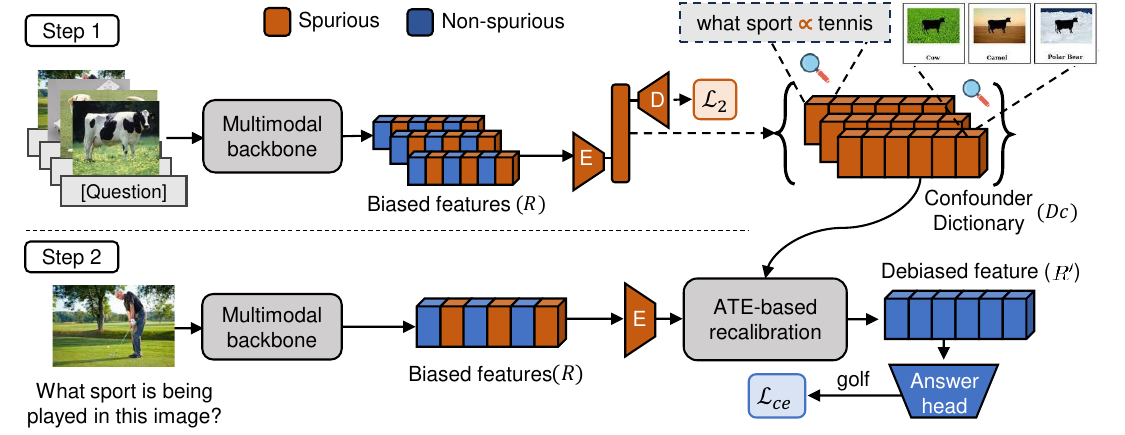}
\end{center}
\vspace{-10pt}
  \caption{An illustration of our method \ourmethod{} based on autoencoder-based confounder modeling and Average Treatment Effect causal mechanism (see Sec.~\ref{sec:deconf}). The confounders are modeled using autoencoder in Step 1 and biased features are recalibrated using confounders to get debiased features in Step 2.}
  \vspace{-10pt}
\label{fig:ate}
\end{figure*}

\paragraph{Average Treatment Effect.} The aim of causal inference is to estimate the independent effect of an intervention on a treatment variable $M$ on an outcome of interest $A$ i.e. to estimate the conditional probability distribution $P(A|do(M))$ where the \textit{do}-operation implies the causal effect of $M\rightarrow A$. However, standard models are optimized to infer the observational conditional probability $P(A|M)$. In the presence of confounders i.e. variables $c \in C$ that affect both $A$ and $M$, $P(A|M) \neq P(A|do(M))$. $P(A|do(M))$ can be estimated using backdoor adjustment by controlling for all values of the confounders $c \in C$ as:
\begin{equation}
    P(A|do(M)) = E_{c\sim C}[P(A|M,c)]
    \label{eqn:ate}
\end{equation}
This translates to an empirical sum over all possible values of the confounder in practice, also known as the average treatment effect (ATE) (see Fig.~\ref{fig:causal_graph}(b)). When the confounders are known and observed, the confounder values are selected using suitable heuristics \cite{pearl2000models}. However, observing all confounders is not always possible. Hence, in our instantiation of ATE, we model the variables that can be used as substitutes for the confounders via latent representations in autoencoders \cite{sen2017contextual, kallus2018causal}. \citet{huang2022deconfounded} use average treatment effect-based debiasing for the task of visual grounding by modeling confounders.

\paragraph{Total Effect.} 
We need to isolate the causal effect of $M=m$ on $A$, free from the influence of the confounders $C$. According to causal theory, the total effect (TE) of treatment $M=m$ on $A$ is,

\begin{equation}
    TE = A_{m, C_{m}} - A_{m*, C_{m}}
\end{equation}

where $M=m$, $M=m*$ represent `treatment' and `no treatment' conditions, respectively; $C_{m}$ is the confounder under treatment, and $A_{m, C_{m}}$ is the answer in the presence of treatment as well as confounder. The direct effect of $C_{m}$ on $M$ is eliminated by retaining the confounder on both sides of the difference (see Fig.~\ref{fig:causal_graph}(c)). In our implementation of TE, we take the difference between feature representations of $A_{m, C_{m}}$, $A_{m*, C_{m}}$ i.e. $Z_{m, c}$, $Z_{m*,c}$ respectively, to eliminate the effect of $C_{m}$ (see Sec.~\ref{sec:ratedistort}).

\begin{figure*}[ht]
\begin{center}
  \includegraphics[width=0.85\linewidth]{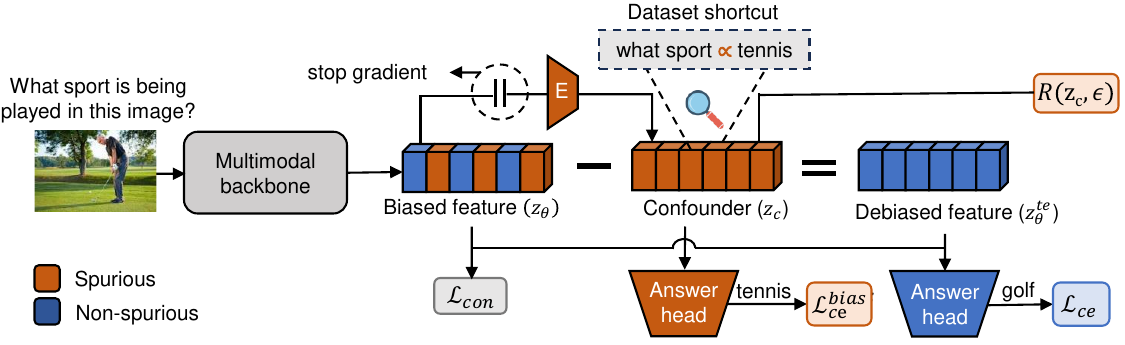}
\end{center}
\vspace{-10pt}
  \caption{An illustration of our method \ratedistort{} based on Rate-Distortion \& Total Effect causal mechanism (see Sec.~\ref{sec:ratedistort}). The biased features are used to learn confounder features guided by rate-distortion minimization and cross-entropy loss ($L_{ce}$). The confounders are subtracted from the biased features to get debiased features.}
  \vspace{-10pt}
\label{fig:rate}
\end{figure*}
\section{Debiasing Methods: \ourmethod{} and \ratedistort{}}

\citet{kirichenko2022last} show that machine-learning models learn spurious as well as non-spurious features when trained on a biased dataset, but over-rely on the former for making predictions. In Sec.~\ref{sec:intro}, we discussed how confounder variables contribute to these spurious predictions. Further, \citet{yang2022chroma} show empirically that deep models preferentially encode dataset shortcuts under limited representation capacity. Indeed, neural nets are expected to trade-off between maximal compression of the learnt representations and maximal fitting to the labels (Information-Bottleneck) \cite{shwartz2022opening}. Hence, we propose information minimization, by limiting representation capacity via low-dimensional vectors, to learn the bias/confounder features. Similar approaches exist i.e. \citet{kallus2018causal} recover latent confounders by performing low-rank matrix factorization on high-dimensional data, and \citet{sen2017contextual} use a low-dimensional variable to encode confounder. We propose two methods to learn and use confounder features for debiasing: (a) latent variable modeling in \ourmethod{} and (b) rate-distortion minimization in \ratedistort{}. In both approaches, the biased features are projected into low-dimensional vectors through various mechanisms, limiting their representation capacity and promoting information minimization. Sec.~\ref{sec:deconf} and~\ref{sec:ratedistort} further elaborate on these methods. We discuss the advantages of our causal debiasing approaches over data augmentation methods in Sec.~\ref{sec:discussion}.

\subsection{\ourmethod{}: Deconfounding Using Average Treatment Effect }
\label{sec:deconf}

We follow a 2-step framework where we start with a pre-trained biased model, then (1) obtain the substitute confounders from the latent variables of autoencoder \cite{huang2022deconfounded} and (2) use these confounders to debias the pretrained model using feature reweighing \cite{kirichenko2022last}.

\paragraph{Step 1:} We collect the biased features $r\in R$ from a biased model for all samples in the training data and train an autoencoder composed of dense layers ($F_{enc}, F_{dec}$) to encode them into a lower dimension (see top, Fig.~\ref{fig:ate}). The latent dimensions of the generative model capture the most common biases in the dataset and serve as a substitute for the confounders. We use a small-capacity network in order to capture the biases stemming from spurious correlations in the latent dimensions and avoid encoding the correct predictive features. $F_{enc}, F_{dec}$ are trained using the reconstruction loss $L_{recon} = d(R,R)$, where $d(,)$ is the Euclidean distance function. We model the substitute confounders $\hat{c}\in \hat{C}$ for R ($\hat{.}$ represents approximation) and cluster them to get a dictionary $D_{\hat{c}}$, which represents the main elements of $\hat{C}$ for efficient backdoor adjustment (Eqn.~\ref{eqn:ate}).

\paragraph{Step 2:} 

\citet{kirichenko2022last} show that non-spurious features can be emphasized in biased features by reweighing them using a balanced dataset. However, creating balanced data is non-trivial for complex tasks like VQA. To overcome this challenge, we instead create an instantiation of backdoor adjustment that reweighs biased features based on their similarity with the substitute confounders (see bottom, Fig.~\ref{fig:ate}). We hypothesize that this leads to lower weights for the simple spurious features and higher weights for more complex predictive features, alleviating the over-reliance on spurious features for prediction. For a sequence of biased features $r = [r_1, r_2, ..., r_k]$, we recalibrate each $r_i$ according to their similarity with the confounders in $D_{c}$ i.e., the weight $w_i$ for $r_i$ is, 
\begin{equation}
    w_i = 1-\frac{1}{len(D_{\hat{g}})}\sum_{g_j \in D_{\hat{g}}} s(F_{enc}(r_i), g_j)
\end{equation}
where $s(.)$ is the cosine-similarity function (see ATE-based recalibration in Fig.~\ref{fig:ate} and see Appendix for an explanation of recalibration as an instantiation of back-door adjustment). 

\begin{equation}
    r'_i = w_i*r_i; R' = [r'_1, r'_2, ..., r'_k]
\end{equation}

The resulting debiased features $R'$ are then used to replace $R$ as shown in Fig.~\ref{fig:ate}.

\subsection{\ratedistort{}: Debiasing Using Rate-Distortion \&  Total Effect}

\label{sec:ratedistort}

The rate-distortion function $R(Z, \epsilon)$ measures the minimum number of bits per vector required to encode the sequence $Z = \{z_1, z_2, ...z_n\} \in \mathcal{R}^{n\times d}$ such that the decoded vectors $\{\hat{z}\}_{i=1}^n$ can be recovered up to a precision $\epsilon^2$ i.e.,
\begin{equation}
    R(Z, \epsilon) = \frac{1}{2}\textrm{log}_{2}\textrm{det}(I + \frac{d}{n\epsilon^2}ZZ^T)
\end{equation}
where $\frac{1}{n}ZZ^T$ is the estimate of covariance matrix for the Gaussian distribution \cite{chowdhury-chaturvedi-2022-learning} and assuming that the vectors are i.i.d. samples from $\mathcal{N}(0,1)$. Rate-distortion values are higher for distribution with high variance (diverse features). Hence, we minimize the rate-distortion to learn confounder representations in \ratedistort{}. Our implementation is illustrated in Fig.~\ref{fig:rate}. Given a \textit{biased model} with parameters $\theta$, we first obtain the biased feature $z_{\theta}$. Then, we encode the $z_{\theta}$ into a lower dimension to promote information loss, along with a classification head ($\mathcal{L}_{ce}^{conf}$) to encourage retaining predictiveness of the information present in the encodings, which we treat as the confounder representation $z_c$. Finally, we enforce rate-distortion minimization ($R(z_{c}, \epsilon)$) on $z_{c}$ for promoting the loss of complex feature information. We enforce a stop gradient (see in Fig.~\ref{fig:rate}) prior to the encoder in order to prevent the training signals for learning confounder representations from seeping into the parameters of the biased model.

In order to isolate the causal effect of $M$, we need to cut off the link $C\rightarrow M$ (see Fig.~\ref{fig:causal_graph}(c)). This can be achieved by computing the total effect (see Sec.~\ref{sec:prelim}) i.e., $A_{m, c} - A_{m*, c}$, where $m$ and $m*$ represent the treatment and no-treatment conditions respectively, while $c$ represents the confounder resulting from $M=m$. We implement this at the feature level by representing $A_{m,c}$ with the biased features $z_\theta$ and $A_{m*,c}$ with the confounder features $z_c$. Next, we take the difference of those features to secure $z_{\theta}^{te}$, which represents the direct effect of $M$. i.e. $z_{\theta}^{te} = z_{\theta} - z_c$. We further aid the debiasing process by enforcing a contrastive loss between the three sets of features $z_{\theta}, z_{c}, z_{\theta}^{te}$ as:
\begin{equation}
    \mathcal{L}_{con} = \textrm{log}\frac{\mathbf{e}^{s(z_{\theta}^{te}, z_{\theta})}}{\mathbf{e}^{s(z_{\theta}^{te}, z_{\theta})} + \mathbf{e}^{s(z_{\theta}^{te}, z_c)}}
\end{equation}
where $s(.)$ is the cosine similarity function. The contrastive loss penalizes the model when the confounder is correlated with the biased feature $z_{\theta}$ and, hence, promotes the debiasing of the multimodal backbone itself. In summary, we jointly optimize the model for learning confounder representations via $\mathcal{L}_{ce}^{conf}, R(Z_{c}, \epsilon)$ and debiasing with the help of the learned confounders via $\mathcal{L}_{con}, \mathcal{L}_{ce}$ i.e., $\theta_{deconf} = \textrm{argmin}_{\theta} \mathcal{L}_{con} + \mathcal{L}_{ce} + \mathcal{L}_{ce}^{conf} + \alpha R(Z_{c}, \epsilon)$, where $\alpha$ is the weight factor for rate-distortion loss.

\subsection{Causal Debiasing vs. Data Augmentation}
\label{sec:discussion}
Data augmentation is an effective and popular method for enhancing model robustness \citep{puli2023nuisances, gokhale-etal-2020-mutant, chen2020counterfactual}, however, it presents certain constraints, particularly when employed in the context of debiasing within VQA models, such as:

\paragraph{Dependency on prior knowledge. } Data augmentation typically hinges on pre-existing knowledge of potential biases within the dataset. For instance, \citet{mikołajczykbareła2023data} use knowledge of biases i.e. the presence of shape and texture bias in data, to augment data based on style transfer, \citet{gokhale-etal-2020-mutant} identify unimodal biases to augment multimodal datasets. However, such awareness may not be comprehensive or entirely precise. Consequently, the efficacy of data augmentation is contingent on the accuracy and completeness of the a priori understanding of the biases underpinning the augmentation strategy. Conversely, methods that manipulate representation vectors directly to remove biases, such as our proposed debiasing techniques, extract spurious correlations from the data without requiring predefined assumptions about specific biases.

\paragraph{Scalability and cost implications.} The creation of augmented datasets is often time-intensive as well as cost-intensive \cite{sauer2020counterfactual}. The process demands domain expertise to adeptly identify and apply augmentations \cite{tang2020onlineaugment}. This resource-intensive nature of data augmentation can curtail its applicability, especially when used for models that must adapt to a multitude of diverse, evolving sources of bias.\\

Automated discovery of spurious correlations, as performed in our proposed methods \ourmethod{} and \ratedistort{}, is advantageous over data augmentation when dataset biases are inadequately defined or in a state of perpetual flux. For instance, in numerous \textit{real-world applications}, the dataset may harbor concealed or subtle biases that evade detection through manual inspection or domain expertise. Similarly, in \textit{dynamic environments}, dataset biases can undergo periodic shifts. As a result, pre-established augmentation strategies become unviable for such scenarios. The techniques proposed in this work can adapt to the changing characteristics of data within a black box, making them more useful.\\

Another research thread aims to uncover coherent data subsets on which machine learning models exhibit subpar performance, such as the approach introduced in Domino \cite{eyuboglu2021domino}. When these underperforming slices are accurately identified and labeled, it offers an opportunity to enhance model robustness by either updating the training dataset or employing optimization techniques designed to handle systematic performance issues in these slices. While this method aligns with our objective of improving the identification of systematic biases, slice discovery approaches achieve it from a data perspective and require ground truth labels, whereas we take a distinct feature-based approach that does not rely on the ground truth.

\section{Measuring Sufficiency \& Necessity of Spurious Features in Multimodal Tasks}
\label{sec:analysis}

\begin{figure}[h]
\vspace{-5pt}
\includegraphics[width=0.43\textwidth]{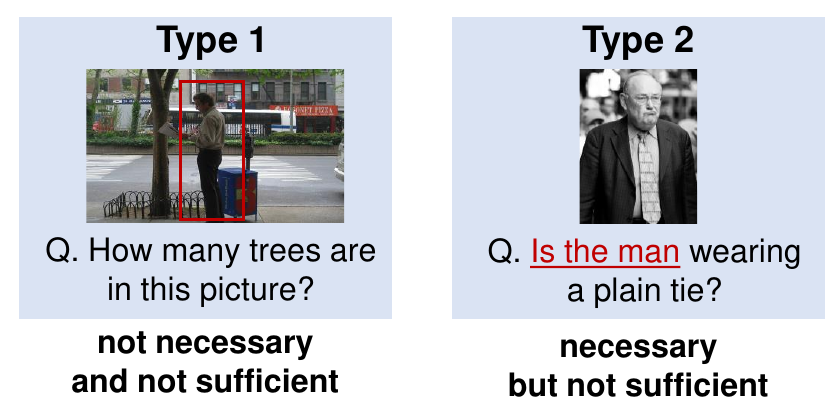}
\caption{Types of spurious features (red) in VQA based on necessity and sufficiency.}
\vspace{-5pt}
\label{fig:matrix}
\end{figure}

OOD generalization accuracies indicate the model's ability to learn causal relationships between inputs and labels \cite{veitch2021counterfactual}. Another approach to assess causal learning is by examining the models' invariance to spurious features in the dataset. \citet{joshi2022all} categorize spurious features into (a) \textit{Type 1 Features} that are neither necessary nor sufficient for predicting the label e.g., `person' (visual feature) when the VQA question is ``How many trees are in the picture?'' (see left, Fig.~\ref{fig:matrix}) and (b) \textit{Type 2 Features} that are necessary but not sufficient to make predictions e.g., the feature ``Is the man'' (see right, Fig.~\ref{fig:matrix}). When a model consistently answers "yes" to all "Is the man..." questions regardless of the image, it is considered to exhibit spurious behavior. We employ this framework to analyze debiasing methods in our experiments.

\paragraph{Necessity.}
To assess the robustness of models to Type 1 features, we compare their performance on samples with and without a specific Type 1 feature. In an unbiased model, the absence of this feature should have no impact on performance. However, a biased model tends to rely on it due to spurious correlations that confound the features and labels. An effective debiasing method should render the model invariant to such features. Type 1 features predominantly arise from the image in multimodal tasks, as depicted in Fig.~\ref{fig:matrix}. Therefore, we evaluate the necessity of these features using counterfactual images \cite{agarwal2020towards} (refer to Sec.\ref{sec:exp}).

\paragraph{Sufficiency.}
To assess the robustness of models to Type 2 features, we propose a new metric for measuring the sufficiency of a feature in relation to a prediction. The certainty of predictions is determined by the Kullback-Leibler (KL) divergence between the predicted output distribution and a uniform distribution across all samples in the group \cite{yingvisfis}. We define the sufficiency score ($\lambda$) as the percentage of the model's certainty that can be attributed to the spurious component of the input in making a prediction. For a data sample $(x, y)$, where the input $x$ consists of the spurious feature $x^{s}$ and the remaining context $x^{c}$, i.e., $x = [x^{s}; x^{c}]$, we compute the sufficiency $\lambda$ as:

\begin{equation}
\label{eqn:metric}
\lambda = \frac{\sum_{i=1}^{G}\textrm{KL}(f(y_{i}|x_{i}^{s})||\mathbf{U})}{\sum_{i=1}^{G}\textrm{KL}(f(y_{i}|x_{i})||\mathbf{U})}
\end{equation}

Here, $\mathbf{U}(.)$ represents the uniform distribution, $f(.)$ denotes the trained model, and $G$ is a group of samples. A reliable debiasing technique should reduce the sufficiency of spurious features. In the case of the multimodal Visual Question Answering (VQA) task, where $x_{i} = (q_{i}, v_{i})$, we evaluate the sufficiency of Type 2 features that arise in the textual modality $q_{i}$. To compute $f(y_{i}|q_{i}^{s}, v_{i})$, we mask $q_{i}^{c}$ in the query before feeding it as input to $f(.)$.

\begin{table*}[!ht]
    \centering
    \small
    \begin{adjustbox}{max width=0.95\textwidth,center}
    \begin{tabular}{l l l l l l l l l r}
    \hline
        ~& \multicolumn{4}{c}{\bf VQA-CP} & \multicolumn{4}{c}{\bf IVQA-CP} & Additional \\ \cline{2-5}\cline{6-9}
        ~ & Overall & Yes/No & Num & other & Overall & Yes/No & Num & other & \#MFLOPS\\ \hline
        LXMERT \cite{tan-bansal-2019-lxmert} & 41.2 & 44.1 & 13.9 & 47.2 & 35.0 & 43.3 & 12.7 & 36.8 & -\\ 
        + IRM \cite{peyrard-etal-2022-invariant} & 42.7 & 44.1 & 15.2 & 49.5 & 36.5 & 43.2 & 12.8 & 39.3 & -\\
        + \ourmethod{} (ours) & 42.2 & 43.6 & 14.6 & 49.0 & 35.8 & 42.9 & 13.2 & 38.2 & {\bf 0.7}\\ 
        + \ratedistort{} (ours) & 43.4 & \underline{48.3} & 14.4 & 48.8 & 36.7 & \underline{46.5} & 12.8 & 38.1 & 8.8\\
        + CD-VQA \cite{kolling2022efficient} & 42.1 & 42.7	& 14.8	& 49.3 & 36.3 & 44.7 & 12.9 & 38.7 & -\\
        + GenB \cite{cho2023genb} & \textbf{52.8} & \textbf{67.3}	& \textbf{29.8}	& \underline{49.7} & \textbf{41.3} & \textbf{50.7} & \textbf{16.7} & \textbf{39.4} & 50.2\\
        D-VQA$_{f}$ \cite{wen2021debiased} & \underline{43.9} & 47.5 & \underline{15.7} & \textbf{49.8} & \underline{37.3} & 45.8 & \underline{13.9} & \underline{39.2} & 18.9\\
        \hline
        D-VQA$_f$ + \ourmethod{} & 43.9 & 47.2 & \textbf{15.9} & 49.9  & 37.4 & 45.7 & 13.9 & 39.3 & 19.6\\ 
        D-VQA$_f$ + \ratedistort{} & {\bf 44.6} & \textbf{47.8} & 15.7 & \textbf{50.8} & \textbf{37.8} & \textbf{46.2} & 13.9 & \textbf{40.1} & 27.7\\ 
        \hline 
        D-VQA  & 52.4 & 65.5 & 29.7 & 51.8 & 44.6 & 62.9 & 26.4 & 39.9 & 25.0\\ \hline
    \end{tabular}
    \end{adjustbox}
    \vspace{-5pt}
    \caption{Accuracy results on the VQA-CP \cite{vqa-cp} and IVQA-CP \cite{agarwal2020towards} test sets. Higher is better. Column `Additional MFLOPs' represents extra MFLOPS introduced by each method over the LXMERT backbone. We report results using a LXMERT model free of the data leakage issue.}
    \label{tab:main}
\end{table*}

\section{Experiment Setup}

\label{sec:exp}
\paragraph{Datasets.}
We evaluate the performance of our methods in both in-distribution (ID) and out-of-distribution (OOD) settings on multiple multimodal tasks, including VQA-CP \cite{agrawal2018don}, GQA \cite{hudson2019gqa}, GQA-OOD \cite{kervadec2021roses}, and NLVR2 \cite{suhr2019corpus} datasets. To further assess robustness in the presence of language and vision biases, we create the \textbf{IVQA-CP} test set by replacing the original images in the VQA-CP test set with counterfactual images from IV-VQA \cite{agarwal2020towards}. These IV-VQA images have been edited to remove irrelevant objects while preserving the original ground truth label (details in Appendix).

\paragraph{Architectures.} We use the LXMERT \cite{tan-bansal-2019-lxmert} model as our baseline and implement our methods \ratedistort{} and \ourmethod{} on top of LXMERT for all datasets. Since VQA-CP is a re-organization of the VQA v2 dataset and LXMERT is pretrained on VQA v2, initializing the pretrained LXMERT model for finetuning on VQA-CP leads to data leakage and an unreasonable increase in accuracy. Hence, we train LXMERT-based models, and baselines from scratch in our experiments and are not comparable to numbers in \citet{wen2021debiased, gokhale-etal-2020-mutant} affected by data leakage.

\section{Results \& Discussion}

\begin{figure}[t]
    \centering
    \includegraphics[width=0.45\textwidth]{
    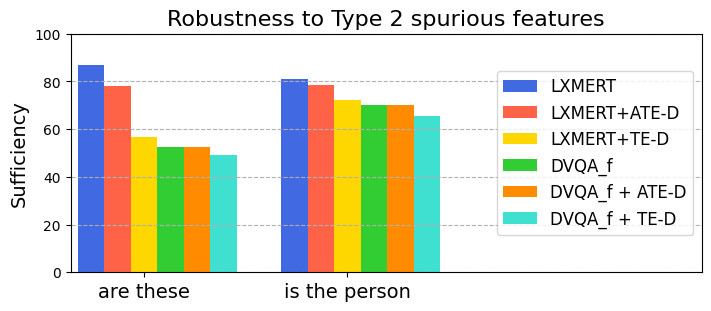}
    \vspace{-5pt}
    \caption{Using our sufficiency metric ($\lambda$, lower is better), we show that our debiased models rely less on Type 2 spurious features than baseline models.
    }
    \vspace{-5pt}
    \label{fig:suff}
\end{figure}
In this section, we discuss the results from the evaluation of our methods for generalization, robustness, effectiveness, and efficiency, and analysis of the learned confounder representations.

\subsection{Does causal debiasing help improve out-of-distribution generalization?}
\label{sec:ood}
 
\begin{figure}[t]
    \centering
    \includegraphics[width=0.45\textwidth]{
    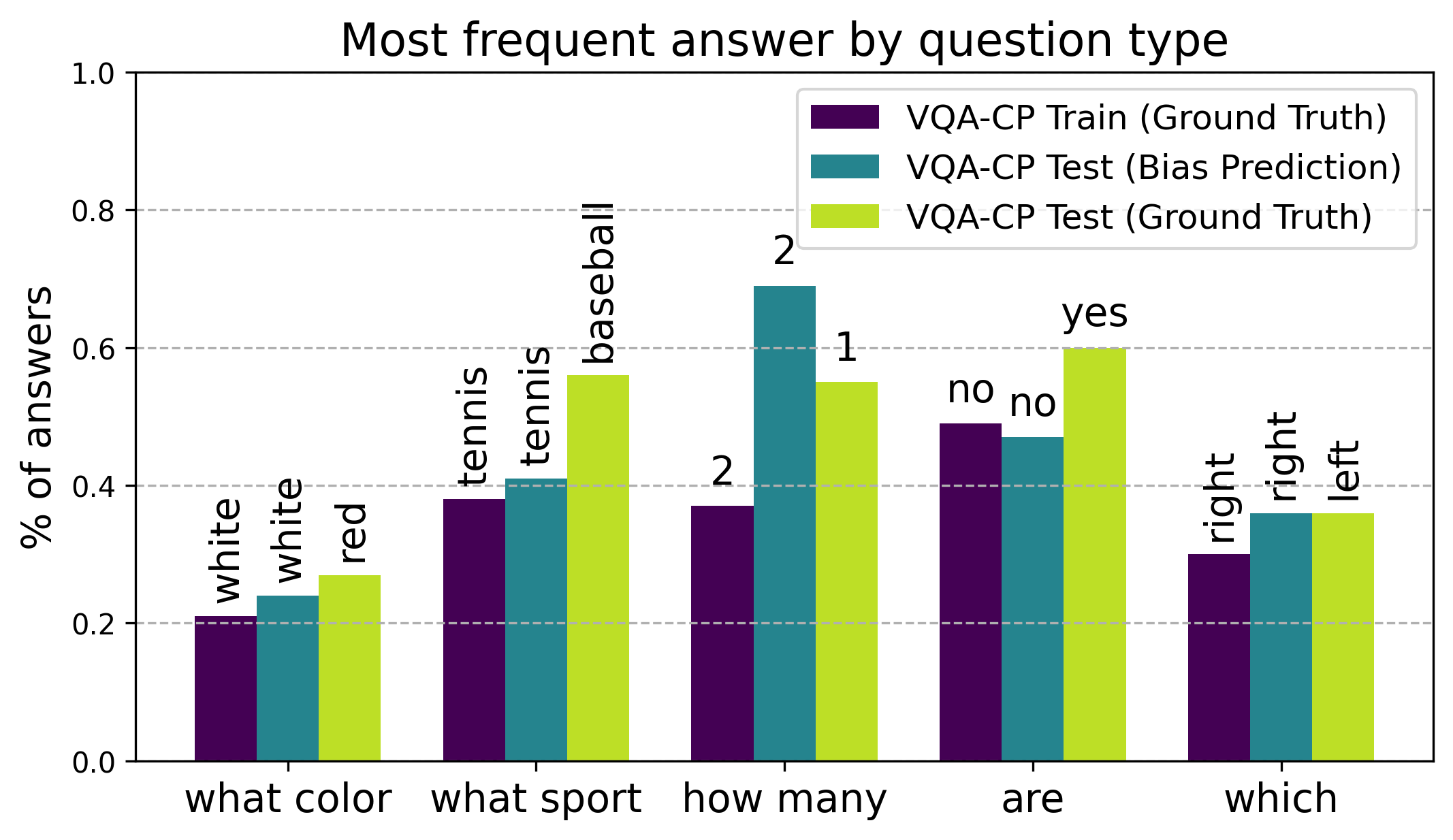}
    \vspace{-5pt}
    \caption{Most frequent answer by question type in VQA-CP train, test, and bias predictions from \ratedistort{}.}    
    \label{fig:bias}
    \vspace{-10pt}
\end{figure}

We evaluate the effect of causal debiasing on improving generalization by evaluating our methods on three multimodal datasets. First, we observe that our methods, \ourmethod{} and \ratedistort{}, demonstrate 1\% and 2.2\% gains over LXMERT on the VQA-CP test set (see Tab.~\ref{tab:main}). \ratedistort{} improves the accuracy of Yes/No category by 4.2\% which has higher bias presence as seen in Fig.~\ref{fig:bias} and outperforms D-VQA$_{f}$, a state-of-art unimodal debiasing method for VQA (feature perspective only), by 0.8\% ($p$=0.04) \footnote{Statistical significance is computed with 100K samples using bootstrap \cite{noreen1989computer, tibshirani1993introduction}. All other gains are statistically significant.} in the Yes/No category, while the latter achieves better overall accuracy on VQA-CP. However, our methods can be used to debias features in any backbone and task, in contrast to D-VQA$_{f}$ that has been designed for VQA. Moreover, D-VQA$_f$ trains a debiased model from scratch while \ratedistort{} debiases a biased model with a few epochs of fine-tuning (see efficiency in Sec.~\ref{sec:cross-modal}). GenB \cite{cho2023genb} achieves state-of-the-art results on top of LXMERT by using ensembles of distilled models but compromises on efficiency. We see 1.8\% and 2.3\% gains in GQA-OOD accuracy with \ourmethod{} and \ratedistort{} over the LXMERT baseline (see Tab.~\ref{tab:gqa}). The GQA-OOD dataset is further divided into OOD-Head and OOD-Tail splits, which represent the samples containing answers from the head and tail of the answer distributions, respectively; our methods achieve improvements in both groups. These gains are obtained along with gains in in-distribution (ID) accuracy on GQA (see Tab.~\ref{tab:gqa}). Additionally, we see 0.4\%, 0.5\% gains with \ourmethod{}, \ratedistort{} respectively on NLVR2, an ID evaluation setting for visual entailment task (see Tab.~\ref{tab:nlvr}). This shows that our methods do not hurt in-distribution performance and are task-agnostic.

\subsection{What kind of biases are captured by confounder representations?}
    
\paragraph{ATE-D.} First, we find that up-weighting features similar to the confounders learned in \ourmethod{}, as opposed to down-weighting (see Sec.~\ref{sec:deconf}), significantly hurts OOD accuracy, implying that the confounder representations indeed encode spurious correlations. Next, we train a non-linear probe on the confounder representations for the VQA task. The accuracy of this probe is 25\%, and the distribution of predicted answers of this probe has lower entropy than that of the predicted answer distribution from unbiased features. Lower entropy suggests higher bias in the semantic concepts encoded in the confounders.

\paragraph{TE-D.} 
The bias representations in \ratedistort{} capture the most prominent input-output biases in the VQA-CP train set, accounting for answers in 0.34\% of the answer vocabulary but covering approximately 67\% of the train questions. The classifier head connected to these bias representations achieves 28\% accuracy on the VQA-CP test set, while the overall causal model accuracy is 44\%. The most frequent answers predicted by this classifier head on the VQA-CP test set align with those in the VQA-CP train set, showing that the captured confounders effectively represent dataset biases (see Fig.~\ref{fig:bias}).

\subsection{Does causal debiasing improve robustness to spurious features?}
\label{sec:robust}

\paragraph{Type 1 Spurious Features.} 
In Sec.~\ref{sec:analysis}, we discuss Type 1 spurious features that are irrelevant to the target output. Our IVQA-CP test set (Sec.~\ref{sec:exp}) shares question annotations with VQA-CP but has images edited to remove irrelevant objects \cite{agarwal2020towards}. Models trained on VQA-CP are evaluated on this dataset, allowing assessment of their robustness to spurious features. The LXMERT baseline shows a significant drop from 41.2\% to 35.0\% on IVQA-CP (Tab.~\ref{tab:main}), indicating the evaluation's challenging nature. Our methods, \ourmethod{} and \ratedistort{}, achieve 0.8\% and 1.7\% improvements respectively over LXMERT on the IVQA-CP test set, enhancing robustness to Type 1 features. D-VQA$_f$ performs explicit visual debiasing and, hence, exhibits the highest robustness to Type 1 features in IVQA-CP.

\paragraph{Type 2 Spurious Features.}
\label{type2}

A prominent source of Type 2 spurious features in VQA is the first few words of a question, as seen in Fig.~\ref{fig:matrix}. We introduce the sufficiency score ($\lambda$, see Eqn.~\ref{eqn:metric}) to understand whether debiasing methods truly improve the robustness of models to such spurious features. We select two question types i.e. questions starting with ``Are these'' and ``Is this person'', which are strongly biased in the training set of VQA-CP, and compute the sufficiency of the phrases for model predictions by masking the remaining question (see Sec~\ref{sec:analysis}). As shown in Fig.~\ref{fig:suff}, we find that causal debiasing methods lower the sufficiency score of the spurious feature for both of these question types, suggesting that they indeed alleviate the reliance of these models on spurious features for making predictions. \ratedistort{} and D-VQA$_f$ achieve similar sufficiency scores, suggesting that they are equally effective at improving robustness by giving more importance to the context. \ratedistort{} achieves lower $\lambda$ than \ourmethod{} which aligns with its larger accuracy gains (see Tab.~\ref{tab:main}). 

\begin{table}[!t]
    \centering
    \small
    \begin{adjustbox}{max width=0.48\textwidth,center}
    \begin{tabular}{l l l l l}
    \hline
        ~ & {\bf GQA} & \multicolumn{3}{c}{\bf GQA OOD} \\ \cline{3-5}
        ~ & ID & Tail & Head & All \\ \cline{2-5}
        LXMERT \cite{tan-bansal-2019-lxmert} & 59.8 & 49.8 & 57.7 & 54.6 \\ 
        +VILLA \cite{gan2020large} & - & 49.9 & 57.2 & 54.5 \\ 
        +MANGO \cite{DBLP:journals/corr/abs-2012-08673} & - & - & - & 54.9 \\ 
        +X-CGM \cite{jiang2021x} & - & 49.9 & 57.5 & 55.6 \\ 
        +\ourmethod{} (ours) & {\bf 60.0} & 50.8 & 59.9 & 56.4 \\ 
        +\ratedistort{} (ours) & 59.9 & {\bf 51.4} & {\bf 60.1} & {\bf 56.8} \\ \hline
        
    \end{tabular}
    \end{adjustbox}
    \vspace{-6pt}
     \caption{Accuracy results on GQA ID and OOD datasets for various debiasing methods. Higher is better.}
     \vspace{-7pt}
     \label{tab:gqa}
    
\end{table}

\begin{table}[!t]
    \centering
    \begin{adjustbox}{max width=0.4\textwidth,center}
    \begin{tabular}{ l l l}
    \hline
        ~ & Acc. & Cons. \\ \hline
        LXMERT \cite{tan-bansal-2019-lxmert} & 74.5 & 39.4 \\ 
        +\ourmethod{} (ours) & 74.9 & 39.9 \\
        +\ratedistort{} (ours) & 75.0 & 39.6 \\ \hline
    \end{tabular}
    \end{adjustbox}
    \vspace{-5pt}
    \caption{Accuracy (Acc.) and consistency (Cons.) results on NLVR2 ID test set. Higher is better.}
    \label{tab:nlvr}
    \vspace{-10pt}
\end{table}

\subsection{Is cross-modal debiasing more effective and efficient than unimodal debiasing?}

\label{sec:cross-modal}
D-VQA$_f$ outperforms cross-modal debiasing in Table \ref{tab:main}, but when D-VQA$_f$ is treated as the biased model in \ratedistort{}, additional improvements of 0.7\% ($p$=0.03) are achieved, indicating that cross-modal interactions contribute to bias not addressed by unimodal debiasing. Cross-modal feature-based confounders effectively mitigate biases involving multiple modalities. Our causal debiasing methods demonstrate higher efficiency compared to D-VQA, with \ourmethod{} adding 0.7 MFLOPS and \ratedistort{} adding 3\% additional parameters and 8.8 MFLOPS to LXMERT. In contrast, D-VQA adds 5\% additional parameters and 18.9 MFLOPS during training, requiring more time as it is trained from scratch. Efficiency results for GQA and NLVR are the same as those reported for VQA.

\section{Conclusion}
\vspace{-5pt}
We propose \ourmethod{} and \ratedistort{} to mitigate biases in models by imposing causally-driven information loss on biased features to learn confounders. Experimental results across various multimodal tasks, datasets, and backbones demonstrate that the learned confounders capture biases successfully, and our methods effectively eliminate biases from both unimodal and multimodal interactions.

\section{Limitations}
While we evaluate robustness to spurious features, we do so on specific question types for Type 2 features and specific Type 1 features (irrelevant objects in the image). Getting an all-inclusive robustness metric for evaluating debiasing methods would be insightful.
Approaches that debias using data augmentation or sample balancing, although cumbersome, are more effective than feature-based debiasing approaches, including those proposed in our paper. More analysis is required to understand how the merits of sample-perspective and feature-perspective methods can be merged efficiently.

\section{Broader Impact}
In this work, the biases that we try to mitigate stem from the spurious correlations present in the dataset that lead to a drop in performance in OOD settings. This helps models learn causal associations between inputs and targets and thus brings them closer to real-world deployment as it helps mitigate the unethical use of these models. However, vision-language models may encode other societal stereotypes and biases present in the data they are trained on and also introduce new ones. VL models explored in this paper are not immune to these issues. We are hopeful that our focus on modeling biases and alleviating them is a step towards more inclusive models. 

\section*{Acknowledgments}
We thank Peter Hase, Zhuofan Ying, Jaemin Cho, and Nitish Joshi for their useful insights about this work, and the reviewers of this paper for their helpful feedback. This work was supported by ARO W911NF2110220, DARPA MCS N66001-19-2-4031, ONR N00014-23-1-2356, DARPA ECOLE Program No. HR00112390060. The views, opinions, and/or findings contained in this article are those of the authors and not of the funding agency.

\bibliography{custom}

\begin{thebibliography}{60}
\expandafter\ifx\csname natexlab\endcsname\relax\def\natexlab#1{#1}\fi

\bibitem[{Agarwal et~al.(2020)Agarwal, Shetty, and Fritz}]{agarwal2020towards}
Vedika Agarwal, Rakshith Shetty, and Mario Fritz. 2020.
\newblock Towards causal vqa: Revealing and reducing spurious correlations by invariant and covariant semantic editing.
\newblock In \emph{Proceedings of the IEEE/CVF Conference on Computer Vision and Pattern Recognition}, pages 9690--9698.

\bibitem[{Agrawal et~al.(2016)Agrawal, Batra, and Parikh}]{agrawal-etal-2016-analyzing}
Aishwarya Agrawal, Dhruv Batra, and Devi Parikh. 2016.
\newblock \href {https://doi.org/10.18653/v1/D16-1203} {Analyzing the behavior of visual question answering models}.
\newblock In \emph{Proceedings of the 2016 Conference on Empirical Methods in Natural Language Processing}, pages 1955--1960, Austin, Texas. Association for Computational Linguistics.

\bibitem[{Agrawal et~al.(2018{\natexlab{a}})Agrawal, Batra, Parikh, and Kembhavi}]{vqa-cp}
Aishwarya Agrawal, Dhruv Batra, Devi Parikh, and Aniruddha Kembhavi. 2018{\natexlab{a}}.
\newblock Don't just assume; look and answer: Overcoming priors for visual question answering.
\newblock In \emph{IEEE Conference on Computer Vision and Pattern Recognition (CVPR)}.

\bibitem[{Agrawal et~al.(2018{\natexlab{b}})Agrawal, Batra, Parikh, and Kembhavi}]{agrawal2018don}
Aishwarya Agrawal, Dhruv Batra, Devi Parikh, and Aniruddha Kembhavi. 2018{\natexlab{b}}.
\newblock Don't just assume; look and answer: Overcoming priors for visual question answering.
\newblock In \emph{Proceedings of the IEEE conference on computer vision and pattern recognition}, pages 4971--4980.

\bibitem[{Antol et~al.(2015)Antol, Agrawal, Lu, Mitchell, Batra, Zitnick, and Parikh}]{VQA}
Stanislaw Antol, Aishwarya Agrawal, Jiasen Lu, Margaret Mitchell, Dhruv Batra, C.~Lawrence Zitnick, and Devi Parikh. 2015.
\newblock {VQA}: {V}isual {Q}uestion {A}nswering.
\newblock In \emph{International Conference on Computer Vision (ICCV)}.

\bibitem[{Bahadori and Heckerman(2020)}]{bahadori2020debiasing}
Mohammad~Taha Bahadori and David Heckerman. 2020.
\newblock Debiasing concept-based explanations with causal analysis.
\newblock In \emph{International Conference on Learning Representations}.

\bibitem[{Cadene et~al.(2019)Cadene, Dancette, Cord, Parikh et~al.}]{cadene2019rubi}
Remi Cadene, Corentin Dancette, Matthieu Cord, Devi Parikh, et~al. 2019.
\newblock Rubi: Reducing unimodal biases for visual question answering.
\newblock \emph{Advances in Neural Information Processing Systems}, 32:841--852.

\bibitem[{Chen et~al.(2020)Chen, Yan, Xiao, Zhang, Pu, and Zhuang}]{chen2020counterfactual}
Long Chen, Xin Yan, Jun Xiao, Hanwang Zhang, Shiliang Pu, and Yueting Zhuang. 2020.
\newblock Counterfactual samples synthesizing for robust visual question answering.
\newblock In \emph{CVPR}.

\bibitem[{Chen et~al.(2022)Chen, Zheng, and Xiao}]{DBLP:conf/eccv/0016Z022}
Long Chen, Yuhang Zheng, and Jun Xiao. 2022.
\newblock \href {https://doi.org/10.1007/978-3-031-20059-5\_6} {Rethinking data augmentation for robust visual question answering}.
\newblock In \emph{Computer Vision - {ECCV} 2022 - 17th European Conference, Tel Aviv, Israel, October 23-27, 2022, Proceedings, Part {XXXVI}}, volume 13696 of \emph{Lecture Notes in Computer Science}, pages 95--112. Springer.

\bibitem[{Cho et~al.(2023)Cho, Kim, Ryu, and Kweon}]{cho2023genb}
Jae~Won Cho, Dong-Jin Kim, Hyeonggon Ryu, and In~So Kweon. 2023.
\newblock Generative bias for robust visual question answering.
\newblock In \emph{Proceedings of the IEEE/CVF Conference on Computer Vision and Pattern Recognition (CVPR)}.

\bibitem[{Chowdhury and Chaturvedi(2022)}]{chowdhury-chaturvedi-2022-learning}
Somnath Basu~Roy Chowdhury and Snigdha Chaturvedi. 2022.
\newblock \href {https://doi.org/10.1162/tacl_a_00512} {Learning fair representations via rate-distortion maximization}.
\newblock \emph{Transactions of the Association for Computational Linguistics}, 10:1159--1174.

\bibitem[{Clark et~al.(2019)Clark, Yatskar, and Zettlemoyer}]{clark2019don}
Christopher Clark, Mark Yatskar, and Luke Zettlemoyer. 2019.
\newblock Don’t take the easy way out: Ensemble based methods for avoiding known dataset biases.
\newblock In \emph{Proceedings of the 2019 Conference on Empirical Methods in Natural Language Processing and the 9th International Joint Conference on Natural Language Processing (EMNLP-IJCNLP)}, pages 4069--4082.

\bibitem[{Eyuboglu et~al.(2021)Eyuboglu, Varma, Saab, Delbrouck, Lee-Messer, Dunnmon, Zou, and Re}]{eyuboglu2021domino}
Sabri Eyuboglu, Maya Varma, Khaled~Kamal Saab, Jean-Benoit Delbrouck, Christopher Lee-Messer, Jared Dunnmon, James Zou, and Christopher Re. 2021.
\newblock Domino: Discovering systematic errors with cross-modal embeddings.
\newblock In \emph{International Conference on Learning Representations}.

\bibitem[{Gan et~al.(2020)Gan, Chen, Li, Zhu, Cheng, and Liu}]{gan2020large}
Zhe Gan, Yen-Chun Chen, Linjie Li, Chen Zhu, Yu~Cheng, and Jingjing Liu. 2020.
\newblock Large-scale adversarial training for vision-and-language representation learning.
\newblock \emph{Advances in Neural Information Processing Systems}, 33:6616--6628.

\bibitem[{Geirhos et~al.(2020)Geirhos, Jacobsen, Michaelis, Zemel, Brendel, Bethge, and Wichmann}]{geirhos2020shortcut}
Robert Geirhos, J{\"o}rn-Henrik Jacobsen, Claudio Michaelis, Richard Zemel, Wieland Brendel, Matthias Bethge, and Felix~A Wichmann. 2020.
\newblock Shortcut learning in deep neural networks.
\newblock \emph{Nature Machine Intelligence}, 2(11):665--673.

\bibitem[{Glymour et~al.(2016)Glymour, Pearl, and Jewell}]{glymour2016causal}
Madelyn Glymour, Judea Pearl, and Nicholas~P Jewell. 2016.
\newblock \emph{Causal inference in statistics: A primer}.
\newblock John Wiley \& Sons.

\bibitem[{Gokhale et~al.(2020)Gokhale, Banerjee, Baral, and Yang}]{gokhale-etal-2020-mutant}
Tejas Gokhale, Pratyay Banerjee, Chitta Baral, and Yezhou Yang. 2020.
\newblock \href {https://doi.org/10.18653/v1/2020.emnlp-main.63} {{MUTANT}: A training paradigm for out-of-distribution generalization in visual question answering}.
\newblock In \emph{Proceedings of the 2020 Conference on Empirical Methods in Natural Language Processing (EMNLP)}, pages 878--892, Online. Association for Computational Linguistics.

\bibitem[{Goyal et~al.(2017)Goyal, Khot, Summers-Stay, Batra, and Parikh}]{goyal2017making}
Y.~Goyal, T.~Khot, D.~Summers-Stay, D.~Batra, and D.~Parikh. 2017.
\newblock \href {https://doi.org/10.1109/CVPR.2017.670} {Making the v in vqa matter: Elevating the role of image understanding in visual question answering}.
\newblock In \emph{2017 IEEE Conference on Computer Vision and Pattern Recognition (CVPR)}, pages 6325--6334, Los Alamitos, CA, USA. IEEE Computer Society.

\bibitem[{Huang et~al.(2022)Huang, Qin, Qi, Sun, and Zhang}]{huang2022deconfounded}
Jianqiang Huang, Yu~Qin, Jiaxin Qi, Qianru Sun, and Hanwang Zhang. 2022.
\newblock Deconfounded visual grounding.
\newblock In \emph{Proceedings of the AAAI Conference on Artificial Intelligence}, volume~36, pages 998--1006.

\bibitem[{Hudson and Manning(2019)}]{hudson2019gqa}
Drew~A Hudson and Christopher~D Manning. 2019.
\newblock Gqa: A new dataset for real-world visual reasoning and compositional question answering.
\newblock In \emph{Proceedings of the IEEE/CVF conference on computer vision and pattern recognition}, pages 6700--6709.

\bibitem[{Jabri et~al.(2016)Jabri, Joulin, and van~der Maaten}]{10.1007/978-3-319-46484-8_44}
Allan Jabri, Armand Joulin, and Laurens van~der Maaten. 2016.
\newblock Revisiting visual question answering baselines.
\newblock In \emph{Computer Vision -- ECCV 2016}, pages 727--739, Cham. Springer International Publishing.

\bibitem[{Jiang et~al.(2021)Jiang, Liu, Liu, Nan, and Zheng}]{jiang2021x}
Jingjing Jiang, Ziyi Liu, Yifan Liu, Zhixiong Nan, and Nanning Zheng. 2021.
\newblock X-ggm: Graph generative modeling for out-of-distribution generalization in visual question answering.
\newblock In \emph{Proceedings of the 29th ACM International Conference on Multimedia}, pages 199--208.

\bibitem[{Joshi et~al.(2022)Joshi, Pan, and He}]{joshi2022all}
Nitish Joshi, Xiang Pan, and He~He. 2022.
\newblock Are all spurious features in natural language alike? an analysis through a causal lens.
\newblock In \emph{Proceedings of the 2022 Conference on Empirical Methods in Natural Language Processing}, pages 9804--9817.

\bibitem[{Kallus et~al.(2018)Kallus, Mao, and Udell}]{kallus2018causal}
Nathan Kallus, Xiaojie Mao, and Madeleine Udell. 2018.
\newblock Causal inference with noisy and missing covariates via matrix factorization.
\newblock \emph{Advances in neural information processing systems}, 31.

\bibitem[{Kervadec et~al.(2021)Kervadec, Antipov, Baccouche, and Wolf}]{kervadec2021roses}
Corentin Kervadec, Grigory Antipov, Moez Baccouche, and Christian Wolf. 2021.
\newblock Roses are red, violets are blue... but should vqa expect them to?
\newblock In \emph{Proceedings of the IEEE/CVF Conference on Computer Vision and Pattern Recognition}, pages 2776--2785.

\bibitem[{Kirichenko et~al.(2022)Kirichenko, Izmailov, and Wilson}]{kirichenko2022last}
Polina Kirichenko, Pavel Izmailov, and Andrew~Gordon Wilson. 2022.
\newblock Last layer re-training is sufficient for robustness to spurious correlations.
\newblock In \emph{The Eleventh International Conference on Learning Representations}.

\bibitem[{Kolling et~al.(2022{\natexlab{a}})Kolling, More, Gavenski, Pooch, Parraga, and Barros}]{Kolling_2022_WACV}
Camila Kolling, Martin More, Nathan Gavenski, Eduardo Pooch, Ot\'avio Parraga, and Rodrigo~C. Barros. 2022{\natexlab{a}}.
\newblock Efficient counterfactual debiasing for visual question answering.
\newblock In \emph{Proceedings of the IEEE/CVF Winter Conference on Applications of Computer Vision (WACV)}, pages 3001--3010.

\bibitem[{Kolling et~al.(2022{\natexlab{b}})Kolling, More, Gavenski, Pooch, Parraga, and Barros}]{kolling2022efficient}
Camila Kolling, Martin More, Nathan Gavenski, Eduardo Pooch, Ot{\'a}vio Parraga, and Rodrigo~C Barros. 2022{\natexlab{b}}.
\newblock Efficient counterfactual debiasing for visual question answering.
\newblock In \emph{Proceedings of the IEEE/CVF winter conference on applications of computer vision}, pages 3001--3010.

\bibitem[{Kolling et~al.(2022{\natexlab{c}})Kolling, More, Gavenski, Pooch, Parraga, and Barros}]{9706896}
Camila Kolling, Martin More, Nathan Gavenski, Eduardo Pooch, Otávio Parraga, and Rodrigo~C. Barros. 2022{\natexlab{c}}.
\newblock \href {https://doi.org/10.1109/WACV51458.2022.00263} {Efficient counterfactual debiasing for visual question answering}.
\newblock In \emph{2022 IEEE/CVF Winter Conference on Applications of Computer Vision (WACV)}, pages 2572--2581.

\bibitem[{Li et~al.(2020)Li, Gan, and Liu}]{DBLP:journals/corr/abs-2012-08673}
Linjie Li, Zhe Gan, and Jingjing Liu. 2020.
\newblock \href {https://arxiv.org/abs/2012.08673} {A closer look at the robustness of vision-and-language pre-trained models}.
\newblock \emph{CoRR}, abs/2012.08673.

\bibitem[{Lin et~al.(2022)Lin, Wu, Chen, Li, and Yu}]{lin2022causal}
Xiangru Lin, Ziyi Wu, Guanqi Chen, Guanbin Li, and Yizhou Yu. 2022.
\newblock \href {https://doi.org/10.1609/aaai.v36i2.20052} {A causal debiasing framework for unsupervised salient object detection}.
\newblock In \emph{Proceedings of the AAAI Conference on Artificial Intelligence}, volume~36, pages 1610--1619.

\bibitem[{Liu et~al.(2022)Liu, Liu, Li, Hou, Yu, and Yang}]{liu2022contextual}
Ruyang Liu, Hao Liu, Ge~Li, Haodi Hou, TingHao Yu, and Tao Yang. 2022.
\newblock Contextual debiasing for visual recognition with causal mechanisms.
\newblock In \emph{Proceedings of the IEEE/CVF Conference on Computer Vision and Pattern Recognition}, pages 12755--12765.

\bibitem[{Mikołajczyk-Bareła(2023)}]{mikołajczykbareła2023data}
Agnieszka Mikołajczyk-Bareła. 2023.
\newblock \href {http://arxiv.org/abs/2308.09464} {Data augmentation and explainability for bias discovery and mitigation in deep learning}.

\bibitem[{Niu et~al.(2021)Niu, Tang, Zhang, Lu, Hua, and Wen}]{niu2021counterfactual}
Yulei Niu, Kaihua Tang, Hanwang Zhang, Zhiwu Lu, Xian-Sheng Hua, and Ji-Rong Wen. 2021.
\newblock Counterfactual vqa: A cause-effect look at language bias.
\newblock In \emph{Proceedings of the IEEE/CVF Conference on Computer Vision and Pattern Recognition}, pages 12700--12710.

\bibitem[{Noreen(1989)}]{noreen1989computer}
Eric~W Noreen. 1989.
\newblock \emph{Computer-intensive methods for testing hypotheses}.
\newblock Wiley New York.

\bibitem[{Pan et~al.(2022)Pan, Li, Zhang, and Tang}]{pan2022causal}
Yonghua Pan, Zechao Li, Liyan Zhang, and Jinhui Tang. 2022.
\newblock Causal inference with knowledge distilling and curriculum learning for unbiased vqa.
\newblock \emph{ACM Transactions on Multimedia Computing, Communications, and Applications (TOMM)}, 18(3):1--23.

\bibitem[{Pearl(2022)}]{pearl2022direct}
Judea Pearl. 2022.
\newblock Direct and indirect effects.
\newblock In \emph{Probabilistic and Causal Inference: The Works of Judea Pearl}, pages 373--392.

\bibitem[{Pearl et~al.(2000)}]{pearl2000models}
Judea Pearl et~al. 2000.
\newblock Models, reasoning and inference.
\newblock \emph{Cambridge, UK: CambridgeUniversityPress}, 19(2).

\bibitem[{Peyrard et~al.(2022)Peyrard, Ghotra, Josifoski, Agarwal, Patra, Carignan, Kiciman, Tiwary, and West}]{peyrard-etal-2022-invariant}
Maxime Peyrard, Sarvjeet Ghotra, Martin Josifoski, Vidhan Agarwal, Barun Patra, Dean Carignan, Emre Kiciman, Saurabh Tiwary, and Robert West. 2022.
\newblock \href {https://aclanthology.org/2022.emnlp-main.387} {Invariant language modeling}.
\newblock In \emph{Proceedings of the 2022 Conference on Empirical Methods in Natural Language Processing}, pages 5728--5743, Abu Dhabi, United Arab Emirates. Association for Computational Linguistics.

\bibitem[{Puli et~al.(2023)Puli, Joshi, He, and Ranganath}]{puli2023nuisances}
Aahlad~Manas Puli, Nitish Joshi, He~He, and Rajesh Ranganath. 2023.
\newblock \href {https://openreview.net/forum?id=eZr_xEPesc7} {Nuisances via negativa: Adjusting for spurious correlations via data augmentation}.

\bibitem[{Ramakrishnan et~al.(2018)Ramakrishnan, Agrawal, and Lee}]{ramakrishnan2018overcoming}
Sainandan Ramakrishnan, Aishwarya Agrawal, and Stefan Lee. 2018.
\newblock Overcoming language priors in visual question answering with adversarial regularization.
\newblock \emph{Advances in Neural Information Processing Systems}, 31.

\bibitem[{Sauer and Geiger(2020)}]{sauer2020counterfactual}
Axel Sauer and Andreas Geiger. 2020.
\newblock Counterfactual generative networks.
\newblock In \emph{International Conference on Learning Representations}.

\bibitem[{Selvaraju et~al.(2019)Selvaraju, Lee, Shen, Jin, Ghosh, Heck, Batra, and Parikh}]{selvaraju2019taking}
Ramprasaath~R Selvaraju, Stefan Lee, Yilin Shen, Hongxia Jin, Shalini Ghosh, Larry Heck, Dhruv Batra, and Devi Parikh. 2019.
\newblock Taking a hint: Leveraging explanations to make vision and language models more grounded.
\newblock In \emph{Proceedings of the IEEE/CVF international conference on computer vision}, pages 2591--2600.

\bibitem[{Sen et~al.(2017)Sen, Shanmugam, Kocaoglu, Dimakis, and Shakkottai}]{sen2017contextual}
Rajat Sen, Karthikeyan Shanmugam, Murat Kocaoglu, Alex Dimakis, and Sanjay Shakkottai. 2017.
\newblock Contextual bandits with latent confounders: An nmf approach.
\newblock In \emph{Artificial Intelligence and Statistics}, pages 518--527. PMLR.

\bibitem[{Shannon(1948)}]{shannon1948mathematical}
Claude~Elwood Shannon. 1948.
\newblock A mathematical theory of communication.
\newblock \emph{The Bell system technical journal}, 27(3):379--423.

\bibitem[{Shwartz-Ziv and Tishby(2022)}]{shwartz2022opening}
Ravid Shwartz-Ziv and Naftali Tishby. 2022.
\newblock Opening the black box of deep neural networks via information.
\newblock \emph{Information Flow in Deep Neural Networks}, page~24.

\bibitem[{Suhr et~al.(2019)Suhr, Zhou, Zhang, Zhang, Bai, and Artzi}]{suhr2019corpus}
Alane Suhr, Stephanie Zhou, Ally Zhang, Iris Zhang, Huajun Bai, and Yoav Artzi. 2019.
\newblock A corpus for reasoning about natural language grounded in photographs.
\newblock In \emph{Proceedings of the 57th Annual Meeting of the Association for Computational Linguistics}, pages 6418--6428.

\bibitem[{Tan and Bansal(2019)}]{tan-bansal-2019-lxmert}
Hao Tan and Mohit Bansal. 2019.
\newblock \href {https://doi.org/10.18653/v1/D19-1514} {{LXMERT}: Learning cross-modality encoder representations from transformers}.
\newblock In \emph{Proceedings of the 2019 Conference on Empirical Methods in Natural Language Processing and the 9th International Joint Conference on Natural Language Processing (EMNLP-IJCNLP)}, pages 5100--5111, Hong Kong, China. Association for Computational Linguistics.

\bibitem[{Tang et~al.(2020{\natexlab{a}})Tang, Niu, Huang, Shi, and Zhang}]{tang2020unbiased}
Kaihua Tang, Yulei Niu, Jianqiang Huang, Jiaxin Shi, and Hanwang Zhang. 2020{\natexlab{a}}.
\newblock Unbiased scene graph generation from biased training.
\newblock In \emph{Proceedings of the IEEE/CVF conference on computer vision and pattern recognition}, pages 3716--3725.

\bibitem[{Tang et~al.(2020{\natexlab{b}})Tang, Gao, Karlinsky, Sattigeri, Feris, and Metaxas}]{tang2020onlineaugment}
Zhiqiang Tang, Yunhe Gao, Leonid Karlinsky, Prasanna Sattigeri, Rogerio Feris, and Dimitris Metaxas. 2020{\natexlab{b}}.
\newblock Onlineaugment: Online data augmentation with less domain knowledge.
\newblock In \emph{Computer Vision--ECCV 2020: 16th European Conference, Glasgow, UK, August 23--28, 2020, Proceedings, Part VII 16}, pages 313--329. Springer.

\bibitem[{Tibshirani and Efron(1993)}]{tibshirani1993introduction}
Robert~J Tibshirani and Bradley Efron. 1993.
\newblock An introduction to the bootstrap.
\newblock \emph{Monographs on statistics and applied probability}, 57:1--436.

\bibitem[{VanderWeele(2015)}]{vanderweele2015explanation}
Tyler VanderWeele. 2015.
\newblock \emph{Explanation in causal inference: methods for mediation and interaction}.
\newblock Oxford University Press.

\bibitem[{Veitch et~al.(2021)Veitch, D'Amour, Yadlowsky, and Eisenstein}]{veitch2021counterfactual}
Victor Veitch, Alexander D'Amour, Steve Yadlowsky, and Jacob Eisenstein. 2021.
\newblock Counterfactual invariance to spurious correlations in text classification.
\newblock \emph{Advances in neural information processing systems}, 34:16196--16208.

\bibitem[{Wen et~al.(2021)Wen, Xu, Tan, Wu, and Wu}]{wen2021debiased}
Zhiquan Wen, Guanghui Xu, Mingkui Tan, Qingyao Wu, and Qi~Wu. 2021.
\newblock \href {https://openreview.net/forum?id=Z4ry59PVMq8} {Debiased visual question answering from feature and sample perspectives}.
\newblock In \emph{Advances in Neural Information Processing Systems}.

\bibitem[{Wu and Mooney(2019)}]{wu2019self}
Jialin Wu and Raymond Mooney. 2019.
\newblock Self-critical reasoning for robust visual question answering.
\newblock \emph{Advances in Neural Information Processing Systems}, 32.

\bibitem[{Yang et~al.(2022)Yang, Kirichenko, Goldblum, and Wilson}]{yang2022chroma}
Wanqian Yang, Polina Kirichenko, Micah Goldblum, and Andrew~G Wilson. 2022.
\newblock Chroma-vae: Mitigating shortcut learning with generative classifiers.
\newblock \emph{Advances in Neural Information Processing Systems}, 35:20351--20365.

\bibitem[{Ying et~al.(2022)Ying, Hase, and Bansal}]{yingvisfis}
Zhuofan Ying, Peter Hase, and Mohit Bansal. 2022.
\newblock Visfis: Visual feature importance supervision with right-for-the-right-reason objectives.
\newblock In \emph{Advances in Neural Information Processing Systems}.

\bibitem[{Zhang et~al.(2016{\natexlab{a}})Zhang, Goyal, Summers-Stay, Batra, and Parikh}]{zhang2016yin}
Peng Zhang, Yash Goyal, Douglas Summers-Stay, Dhruv Batra, and Devi Parikh. 2016{\natexlab{a}}.
\newblock \href {https://doi.org/10.1109/CVPR.2016.542} {Yin and yang: Balancing and answering binary visual questions}.
\newblock In \emph{2016 IEEE Conference on Computer Vision and Pattern Recognition (CVPR)}, pages 5014--5022.

\bibitem[{Zhang et~al.(2016{\natexlab{b}})Zhang, Goyal, Summers{-}Stay, Batra, and Parikh}]{DBLP:conf/cvpr/ZhangGSBP16}
Peng Zhang, Yash Goyal, Douglas Summers{-}Stay, Dhruv Batra, and Devi Parikh. 2016{\natexlab{b}}.
\newblock \href {https://doi.org/10.1109/CVPR.2016.542} {Yin and yang: Balancing and answering binary visual questions}.
\newblock In \emph{2016 {IEEE} Conference on Computer Vision and Pattern Recognition, {CVPR} 2016, Las Vegas, NV, USA, June 27-30, 2016}, pages 5014--5022. {IEEE} Computer Society.

\bibitem[{Zhang et~al.(2021)Zhang, Lin, Han, and Sun}]{zhang-etal-2021-de}
Wenkai Zhang, Hongyu Lin, Xianpei Han, and Le~Sun. 2021.
\newblock \href {https://doi.org/10.18653/v1/2021.acl-long.371} {De-biasing distantly supervised named entity recognition via causal intervention}.
\newblock In \emph{Proceedings of the 59th Annual Meeting of the Association for Computational Linguistics and the 11th International Joint Conference on Natural Language Processing (Volume 1: Long Papers)}, pages 4803--4813, Online. Association for Computational Linguistics.

\end{thebibliography}
\bibliographystyle{acl_natbib}

\appendix

\section{Causal Theory Preliminaries}
\label{sec:app_prelim}
In this section, we discuss our proposed causal graph for multimodal tasks and the two causal mechanisms relevant to our debiasing methods.

\paragraph{Causal Graph.} Causal graphs are directed acyclic graphs $\mathcal{G}=\{\mathcal{V}, \mathcal{E}\}$ where the edges $\mathcal{E}$ are used to represent causal relationships between random variables $\mathcal{V}$. An example is shown in Fig.~\ref{fig:causal_graph}(a), where $\mathbf{M}$ has a \textit{direct effect} on $\mathbf{A}$.When the variable $\mathbf{Q}$ has an \textit{indirect effect} on $\mathbf{A}$ through a variable $\mathbf{M}$ i.e. $\mathbf{Q}\rightarrow\mathbf{M}\rightarrow\mathbf{A}$, the variable $\mathbf{M}$ is said to be a \textit{mediator} in the causal graph. If a variable $\mathbf{C}$ has a direct causal effect on both $\mathbf{M}$ and $\mathbf{A}$, it is said to be a \textit{confounder}.

\paragraph{Causal Perspective for Multimodal Tasks.} Models developed for multimodal tasks are designed to use the combined data stream of vision ($V$) and language ($Q$) for solving the task. However, the unimodal data variables may act as confounders and give rise to spurious features in the model e.g. via $Q\rightarrow M, Q\rightarrow A$. Existing approaches that leverage causal theory for debiasing multimodal models aim to eliminate the direct unimodal effects. However, consider the VQA example in Fig.~\ref{fig:intro}. A potential spurious correlation that may lead to incorrect predictions from models on similar examples is that in most training instances where the question asks the color of an object, the object is present in the center of the image. Spurious correlations arising from such multimodal interactions are ignored in existing causal graphs for multimodal tasks. Hence, we propose to model the spurious correlation as a confounder $\mathbf{C}$ that affects the mediator $\mathbf{M}$ and the answer $\mathbf{A}$ (see Fig.~\ref{fig:causal_graph}(a)). This allows us to model the biases encoded in the multimodal features as confounder $\mathbf{C}$ and eliminate the bias using causal intervention.

In order to debias VQA models, we adopt two causal mechanisms i.e., the Average Treatment Effect (ATE) and Total Effect (TE), which essentially refer to the same effect but differ in how they deal with the confounder \cite{vanderweele2015explanation, tang2020unbiased}. In ATE, $C$ is treated as a distribution, and $c$ is sampled without assuming a causal association with the treatment $M=m$. In TE, $c$ is causally associated with the treatment $M=m$ in each sample. We explore both mechanisms in our experiments and discuss their theories below.

\paragraph{Average Treatment Effect.} The aim of causal inference is to estimate the independent effect of an intervention on a treatment variable $M$ on an outcome of interest $A$ i.e. to estimate the conditional probability distribution $P(A|do(M))$. However, standard models are optimized to infer the observational conditional probability $P(A|M)$ and in the presence of confounders i.e. variables $c \in C$ that affect both $A$ and $M$ 
\begin{equation}
    P(A|M) \neq P(A|do(M))
\end{equation}
where the \textit{do}-operation implies the causal effect of $M\rightarrow A$. $P(A|do(M))$ can be estimated using backdoor adjustment by controlling for all values of the confounders $c \in C$, i.e.,
\begin{equation}
    P(A|do(M)) = E_{c\sim C}[P(A|M,c)]
\end{equation}
This translates to an empirical sum over all possible values of the confounder in practice, also known as average treatment effect (ATE) (see Fig.~\ref{fig:causal_graph}(b)). When the confounders are known and observed, the confounder values are selected using suitable rules and heuristics \cite{pearl2000models}.

\paragraph{Total Effect.} 
We need to isolate the causal effect of $M=m$ on $A$, free from the influence of the confounders $C$. According to causal theory, the total effect (TE) of treatment $M=m$ on $A$ can be computed as,
\begin{equation}
    TE = A_{m, C_{m}} - A_{m*, C_{m}}
\end{equation}
where $M=m*$ represents the "no treatment" condition and $C_{m}$ represents the confounder under the treatment condition i.e $M=m$. By retaining the confounder in both sides of the difference, we eliminate the direct effect of $C_{m}$ on $M$ (see Fig.~\ref{fig:causal_graph}(c)).

\subsection{\ourmethod{}}

\label{sec:math}
Step-2 of \ourmethod{}:

Inspired by feature reweighing \cite{kirichenko2022last}, we instantiate backdoor adjustment by recalibrating $r_i$ based on confounder similarity i.e., $E_{\hat{c}\in D_{\hat{c}}}[f(R,\hat{c})]$ (see Fig.~\ref{fig:causal_graph}(b)) as,

\begin{equation}
    P(A|do(Q), do(V)) = P(A|do(M))  
\end{equation}
\begin{equation}
    E_C[P(A|M, C)] = E_{\hat{c}\in D_{\hat{c}}}[P(A|M, \hat{c})]
\end{equation}
\begin{equation}
\approx P(A|E_{\hat{c}\in D_{\hat{c}}}[f(M,\hat{c})])
\end{equation}
See the appendix of \citet{huang2022deconfounded} for complete proof.
In our analysis, we instantiate $f(.)$ as the cosine similarity function in $s(.)$, as discussed in Sec \ref{sec:deconf}.

\section{Analysis}

While OOD generalization accuracies are indicative of the model learning causal relationships between the inputs and labels, another way to probe causal learning is to investigate if the models are robust to spurious features present in the dataset. In order to evaluate this, in this section, we discuss an analysis framework for probing the behavior of models toward spurious features and propose a new metric for evaluation. \citet{joshi2022all} define the probability of necessity (PN)  of a feature $X_i$ for predicting the label $Y$ as the probability that the ground truth label $Y$ changes when the feature $X_i$ is changed. Similarly, they define the probability of sufficiency (PS) of a feature $X_i$ for predicting the label $Y$ as the probability that setting $X_i = x_{i}$ in a sample where $X_i \neq x_i$ is absent changes its ground truth label $Y$. Based on this framework, spurious features are categorized into (a) \textit{low PN, low PS features}: These features are irrelevant to the ground truth label e.g., person in the image when the VQA question is ``How many trees are in the picture?'' (see Fig.~\ref{fig:matrix}) (b) \textit{High PN, low PS features}: These features are necessary but not sufficient to make predictions i.e. the model should rely on other features in their presence. For instance, when a model always answers ``yes'' to all questions starting with ``Is the man..'' irrespective of the image, the model is biased towards the feature ``Is the man..'' (see Fig.~\ref{fig:matrix}). Henceforth, we refer to the low PS, low PS, and high PN, low PS features as \textit{Type 1} and \textit{Type 2} features, respectively. We use this framework to analyze the various debiasing methods in our experiments.

\paragraph{Sufficiency.} In order to evaluate the robustness to \textit{sufficiency} of type 2 features, we propose a novel metric for quantifying the sufficiency of a feature towards a prediction. We define the certainty of predictions as the KL divergence between the predicted output distribution and uniform distribution across all samples in the group \cite{yingvisfis}. We define the sufficiency score ($\lambda$) as the certainty of a model's prediction when only the non-spurious features are the input to the model. Further, in order to make this metric comparable across models, we normalize this with the certainty of the model's predictions when the complete sample i.e., spurious as well as non-spurious features, is the input to the model. This results in a metric that represents the percentage of certainty of the model that can be attributed to the non-spurious component of the input. For a data sample $(x, y)$, let the input $x$ be comprised of the spurious feature $x^{s}$ and the remaining context $x^{c}$ i.e. $x = [x^{s}; x^{c}]$. The sufficiency $\lambda$ is computed as follows:
\begin{equation}   
    \lambda = \frac{\sum_{i=1}^{G}\textrm{KL}(f(y_{i}|x_{i}^{s})||\mathbf{U})}{\sum_{i=1}^{G}\textrm{KL}(f(y_{i}|x_{i})||\mathbf{U})} 
\end{equation}
where $\mathbf{U}(.)$ represents the uniform distribution, $f(.)$ is the trained model, and $G$ is a group of samples. A good debiasing technique should increase the sufficiency of non-spurious features. For the multimodal VQA task where $x_{i} = (q_{i}, v_{i})$, we focus on the type 2 features emerging in the text modality $q_{i}$. To compute $f(y_{i}|q_{i}^{c}, v_{i})$, we mask $q_{i}^{s}$ in the query before sending it as input to $f(.)$.

\section{Experiment Setup}

\begin{table*}[t]
\small
\begin{center}
\def\arraystretch{1.3}
\begin{tabular}{|l|l|l|l|} 
 \hline
 \textbf{Hyperparameter} & \textbf{LXMERT} & \textbf{ATE} & \textbf{TE}  \\
 \hline
 Learning Rate & 5e-5 & 5e-5 & 5e-5\\
 Epochs & 20 & 5 & 5\\
 Max Gradient Norm & 1.0 & 1.0 & 1.0 \\
 Weight Decay & 0.0 & 0.01 & 0.01 \\
 Batch Size & 32 & 32 & 32 \\
 Max Length & 128 & 128 & 128  \\
 Warmup Ratio & 0.1 & 0.1 & 0.1 \\
 LR Decay & Linear & Linear &  Linear \\
 Optimizer & AdamW & AdamW & AdamW \\
Bias dimension factor & - & - & 4 \\
Confounder dictionary size & - & 10 & - \\
 \hline
\end{tabular}
\vspace{-5pt}
\caption{Training hyperparameters for different models trained on the VQA-CP dataset.}
\end{center}
\end{table*}

\subsection{Datasets}
\begin{itemize}
    \item VQA-CP \cite{vqa-cp}: It is a re-organization of the VQAv2 \cite{VQA} such that the distribution of question type-answer correlation is different between the train and test splits. This evaluation helps demonstrate the method's ability to debias in a setting where language bias is dominant.
     
    \item VQA-CP + IV-VQA: We evaluate it on a new version of the VQA-CP test set where we replace the image in each sample with their invariant counterparts from the IV-VQA dataset from \cite{agarwal2020towards}. IV-VQA dataset has images replaced with their edited version obtained after removing irrelevant objects in a way that the predicted answer does not change. This adds another layer of hardness to the benchmark along the image dimension. This evaluation helps demonstrate the method's ability to debias in a setting where both language and vision biases are dominant. 

    \item GQA\cite{hudson2019gqa}, GQA-OOD\cite{kervadec2021roses}:\\ GQA evaluation helps measure visual reasoning as well as compositional question-answering abilities. GQA-OOD is a re-organization of the GQA dataset that introduces distribution shifts in validation and test sets based on question type similar to VQA-CP. 

    \item NLVR2 \cite{suhr2019corpus}: It helps the generalization to multimodal tasks other than question answering. It helps evaluate reasoning abilities about sets of objects, comparisons, and spatial relations. 
\end{itemize}

All our experiments are run with a single seed value.

\paragraph{Baselines.} We use D-VQA$_{f}$ (feature perspective only) \cite{wen2021debiased} based on LXMERT as the baseline for experiments with VQA-CP and train from scratch due to the aforementioned reasons. We also present results from D-VQA (both feature \& sample perspective) for comparison, however, note that methods using data balancing are not comparable to causal debiasing methods (see Sec.~\ref{sec:intro}).

\section{Results}

\subsection{Analysis of confounder features}
We compare the most frequent answer in the VQA-CP training and test sets with those from the predictions of the bias classifier head in \ratedistort{} in Fig.~\ref{fig:bias}. As discussed in Sec.\ref{sec:analysis}, the predictions from bias classifier head closely tracks the distribution of answers in VQA-CP training set, even though the VQA-CP test set distribution is significantly different from VQA-CP train. This shows that the confounder representations indeed capture the strong priors present in the training set.

\paragraph{Explanation and proof for biases stemming from multimodal interactions.} Multimodal models have been known to be brittle to linguistic biases \cite{goyal2017making} and visual biases \cite{wen2021debiased}. In this work, we demonstrate the presence of multimodal biases and the need to remove those biases from multimodal features. (Proof) Many existing debiasing methods focus on removing each unimodal bias (e.g., linguistic) from multimodal features independently of the other unimodal biases (e.g., visual). However, \citet{agarwal2020towards} suggest that the biases can stem from multimodal interactions as well; they perform semantic edits on images in VQA (I-VQA dataset) that should not affect the ground truth, and show that the answers from multimodal models change in response to these invariant edits. (Existing Methods) Indeed, methods like D-VQA \cite{wen2021debiased} leave large room for improvement in terms of performance on the IVQA-CP dataset that are designed to test for multimodal biases, as we show in Table 1. (Our Approach) We formalize this phenomenon through the causal graph proposed in our paper in Fig. \ref{fig:causal_graph}, where we explicitly model the confounders that affect the variable connecting multimodal representation (M) and the outcome (A). The unimodal biases are implicitly modeled via the multimodal variable (Q->M->A, V->M->A). (Example) We demonstrate an example of this phenomenon in Fig \ref{fig:intro}, where D-VQA fails to answer a question from the IVQA-CP test set correctly, and our proposed method, TE-D, is able to answer correctly because of multimodal debiasing. (Empirical Results) Additionally, we show improvements on top of unimodal debiasing methods like DVQA$_f$ with our multimodal debiasing approach (see rows 6,7 in Table \ref{tab:main}). Our goal in this work is to demonstrate the presence of multimodal biases and the need for multimodal debiasing along with the potential of confounder modeling via information loss in causal multimodal debiasing, and our results support this claim.

\end{document}